# Death in Genetic Algorithms

Micah Burkhardt, Roman V. Yampolskiy
Computer Science and Engineering
Speed School of Engineering
University of Louisville, USA
Micah.Burkhardt@louisville.edu, Roman.Yampolskiy@louisville.edu

*Abstract*— **Death has long been overlooked in evolutionary algorithms. Recent research has shown that death (when applied properly) can benefit the overall fitness of a population and can outperform sub-sections of a population that are "immortal" when allowed to evolve together in an environment [1]. In this paper, we strive to experimentally determine whether death is an adapted trait and whether this adaptation can be used to enhance our implementations of conventional genetic algorithms. Using some of the most widely accepted evolutionary death and aging theories, we observed that senescent death (in various forms) can lower the total run-time of genetic algorithms, increase the optimality of a solution, and decrease the variance in an algorithm's performance. We believe that death-enhanced genetic algorithms can accomplish this through their unique ability to backtrack out of and/or avoid getting trapped in local optima altogether.**

## I. INTRODUCTION

"DEATH is not the opposite of life, but an innate part of it." [2]. Evolutionary algorithms have always endeavored to emulate the natural forces of selection that we see in the world around us. While these algorithms have proven to be useful heuristics for optimization, in these algorithms we have largely ignored one of the fundamental "innate" components of life: death.

One of the most essential questions that we have as humans is "Why do we die?". Recent studies [1, 3-6] have brought some light to this question and shown that death may not be just for population control, but may be much more than that. To illustrate this, just as we have evolved with ten fingers and apposable thumbs as this provides us an evolutionary advantage, death may also be a heritable trait that benefits a population and pushes it to be more fit in its environment.

Many theories have been developed on how and why we die. Some of the most popular and widely accepted theories are Programmed Death Theory, Mutation Accumulation Theory, Antagonistic Pleiotropy Theory, Disposable Soma Theory, DNA Damage Theory, and Telomere Shortening Theory. Largely, these theories can be abstracted into three main categories based on how an individual's fitness degrades over time:
1. Rapid Senescence
2. Gradual Senescence
3. "Non-Smooth" Senescence

Rapid senescence [7] can be considered a superset of evolutionary death theories in which a lifeform has a set age at which its fitness "rapidly" deteriorates. This rapid deterioration of its fitness inevitably leads to the lifeform's death. Programmed death is a common member of the rapid senescent theories and is one of the theories tested later in this paper.

Gradual senescence [8] focuses less on when a lifeform dies and more on its fitness as it ages. Gradual senescent theories will state that as a lifeform ages, it becomes less fit for its environment. It has been posited that this aging can be the result of many different factors such as DNA damage, mutation, a lack of resources, or telomere shortening. Alternatively, according to antagonistic pleiotropy, death and aging are the result of genes that provide an advantage earlier in life but become detrimental later. Whatever the cause, gradual senescence focuses on a slow, smooth degradation of fitness over time.

"Non-smooth" senescent theories are not usually given their own category, but for this research, we are assigning them their own classification to differentiate and test them. This superset is unique as lifeforms not only age and degrade over time as in the previous two concepts, but they can repair their degradation to extend their lifespan. This leads to a non-smooth line when the fitness of an individual is graphed as a function of time. However, rapid senescence maintains a constant fitness until death and the fitness of an individual in gradual senescence declines slowly with time. The main theory that will be tested in this category is the Disposable Soma Theory where lifeforms go through periods of repair, growth, and reproduction (favoring reproduction).

Section II of this paper covers recent work in the study of the effects of senescence and death in evolutionary systems. Section III contains our proposed plan for merging evolutionary theories with conventional genetic algorithms and measuring the results. Section IV articulates the results of these experiments. Section V discusses the implications of these findings and where this research may lead.

## II. LITERATURE REVIEW

In 2013, Joshua Mitteldorf and Andre Martin placed automata on a 128 x 128 grid with some automata being allowed to age towards a senescent death and some automata being "non-agers". In this study they showed that "agers" were able to adapt better to the environment than the "non-agers". "…agers prevailed more often than non-agers, increasing their success with decreasing age." Mitteldorf and Martin note that when the programmed death age is too low, the short-lived individuals are always selected against, and if the age is too high, few individuals will live to meet their programmed death (making it have little to no effect). This study shows that with the right aging conditions, death can be a useful tool to guide a population's development. If the age is set too low, then the population is not able to fully exploit its search of an area, and if the age is set too high, the population is not able to fully explore the search space towards optimality [6].

This idea that shorter-lived individuals may win out over individuals with longer or immortal lifespans has been tested in more than a solely virtual environment. In 2016, Kyryakov, Gomez-Perez, and their team of biologists showed that "under laboratory conditions that mimic natural selection within an ecosystem", three mutant strains of Saccharomyces cerevisiae long lifespans are forced out of the ecosystem when placed in an environment with a shorter-lived strain of the same bacteria. Again, this shows that populations of individuals with shorter lifespans can adapt more quickly to an environment than populations with longer-lived individuals as they are more able to rapidly change their genetic material and hence their fitness to better suit their world [9].

In 2013, Werfel, Ingber, and Bar-Yam performed "invasion studies" with cellular automata. In these studies, mortal individuals were introduced into a large population of immortal automata. The automata were then allowed to compete for resources and space or die off. They noted that the "… [mortal individuals] had a success rate typically 2 to 3 orders of magnitude greater than that of immortals, while immortals managed no successful invasions of mortal population in a total of several million trials." Werfel, Ingber, and Bar-Yam went on to say that the results of their experiments show that programmed death and rapid senescence "…are consistent with natural selection" [1].

Before the development of modern evolutionary theory, it was widely believed that death was not possible as an evolved trait as the effects of it were too deleterious and contrary to the individuals' own good. This supposed self-centeredness would initially seem to be of the most benefit to an individual evolutionarily, but Mitteldorf writes: "…altruistic death can be selected in a spatially structured population, only after individuals have already been weakened by senescence". This statement supports the idea that aging and death are intertwined and their correct implementation is vital to a population's success [9].

In 2006, Mitteldorf saw an issue where studies had shown that aging was an adaptive trait, but this was countered by the fact that the benefits gained by aging must be "… too slow and diffuse..." to make up for its cost. Mitteldorf hypothesized instead that aging and death may be an important factor in helping to stabilize the population dynamics preventing population explosion, extinction, and resource depletion. To test this, individuals were placed in a torus and allowed to reproduce locally asexually. Death was controlled by a Gompertz function as well as a function to determine over-crowding in an area. Mitteldorf concluded that when birth-rate is fixed, aging can be used as an adaptation to moderate fluctuations and instability within a population [5].

Theoretical work has also been done on how senescence is handled in conventional evolutionary algorithms. In his 2017 paper on "The Concept of Ageing in Evolutionary Algorithms" [4], Dimopoulos writes of three main categories of selection in evolutionary algorithms: age-based survivor selection, fitness-based survivor selection, and a hybrid survivor selection. In an age-based selection strategy, all chromosomes of a certain age are immediately killed and replaced by an equal number of offspring. Typically, as in a simple genetic algorithm, all parents are replaced at each iteration and are survived by their offspring. The benefit of this approach, Dimopoulos writes, is that it "…reduce[s] the selective pressure applied during the operational steps and hence reduce[s] the probability of the algorithm converging prematurely to a local optimum…". In stark contrast to age-based selection is fitness-based selection in which the selective process does not care if the individual has been around for one iteration or one million. Fitness-based selection will choose the fittest "n" individuals and breed them to replace the "n" least fit individuals. This strategy focuses on increasing the selective pressure as the best performers are maintained until enough individuals with a higher fitness are born. As a single very fit individual tends to produce more offspring in this solution, the genetic diversity of the population can tend to become less dissimilar and converge prematurely. The final selection strategy Dimopoulos discusses is the hybrid selection strategy. The hybrid selection strategy essentially blends the two previous approaches. The strategy is mainly based around an age-based selection, but elitism is implemented in the algorithm to maintain a set number of the fittest members of the population. This approach helps to maintain genetic diversity, but it also helps to ensure that knowledge is not lost from the most elite individuals dying at every iteration. On aging in evolutionary algorithms, Dimopoulos writes: "…the mechanism through which individual solutions 'survive' during the operation of the evolutionary cycle is a significant factor in achieving an equilibrium between the preservation of 'fit' solutions (exploitation) and the systematic development of new ones (exploration)." Here Dimopoulos conjectures that there is an irrefutable link between death and a population's ability to balance the exploration and exploitation of a search space optimally.

III. PROPOSED APPROACH

A. Theory

The three "conventional" evolutionary algorithm approaches that Dimopoulos suggests handle aging very differently than evolutionary aging theories. Age-based selection is similar to the Programmed Death Theory (although the case Dimopoulos suggests wherein all members of the population are replaced at each iteration is an extreme example). Fitness-based selection does not comply at all with any theories of natural aging as individuals in the fittest portion of the population may be "immortal" if their fitness is high enough to never be replaced by an offspring. A hybrid approach, as it combines the two previous approaches, also has the same potential flaw of a fitness-based strategy in that one very fit individual may live and breed forever. This potential for an indefinite lifespan creates a random lifespan for each individual in all but the age-based strategy.

As it has been shown that senescent death can improve the fitness of a population biologically and in cellular automata, we hypothesize that these same senescent techniques can be applied to conventional genetic algorithms to enhance their performance in the same manner. If in fact there is an increase in the genetic algorithms' performances, then this would further show that senescence is an adapted trait for a population. To test this experimentally, we propose that from Dimopoulos's three "conventional" evolutionary algorithm selection techniques, the performance of the best be compared to evolutionary aging/death-enhanced genetic algorithms using rapid senescence (Programmed Death Theory), gradual senescence (Mutation Accumulation/ DNA Damage Theory), and a non-smooth aging function (Disposable Soma Theory). In addition to these experiments, we propose a hybrid cellular automata genetic algorithm (similar to the experiments done by Mitteldorf [5, 6] or Werfel, Ingber, and Bar-Yam [1]) to directly show the effects of death in an aging population compared to a non-aging population while the chromosomes are geographically isolated and allowed to breed locally on a torus.

### B. Test Problem, Genetic Representation, & Constants

To test the theory that senescence is an evolved trait that can be adopted to benefit the performance of conventional genetic algorithms, a hundred-city instance of the symmetric travelling salesman problem [10] was used as a baseline test. One hundred full repetitions of each experiment were run and performance data collected on the same problem instance. Across all experiments, the chromosomes were represented as an object that contained all cities uniquely and randomly ordered into a list to serve as the "genetic material" [11]. In experiments that require aging, the chromosomes were given an attribute to track their current age and maximum age as needed or, in the case of gradual senescence, an age-adjusted fitness metric. Each population is initialized with 30 chromosomes (except for the hybrid cellular automata genetic algorithm "CA+GA" as this is operated on a filled 10 x 10 matrix with edges connected). The chance of a mutation occurring in each gene for every test is kept constant at a one in 10,000 chance. The stopping criteria is also held constant in all experiments, except for the CA+GA, at 20,000 generations that are allowed to pass so that each algorithm may take the exact same number of generations to find its most optimum solution.

### C. Conventional Genetic Algorithms

The group of three genetic algorithms suggested by Dimopoulos [4] will first be tested to find the highest performing of the three. This top-performer will then be used as a comparative base to judge the performance of each of the senescent genetic algorithms.

*a) Age-based Selection*

In the age-based selection conventional algorithm, all chromosomes are immediately replaced by their offspring at each generation. Selection is implemented by allowing the top two performing chromosomes to mate. All other chromosomes are then paired up and allowed to produce offspring. Chromosomes are bred using a random two-point crossover function and then the offspring iteratively replace all chromosomes in the population until none of the original population are left [4].

*b) Fitness-based Selection*

In the fitness-based selection baseline experiment, 60% of the most fit portion of the population is selected at each iteration for reproduction. The top 60% of chromosomes are then paired using the same strategy as the age-based algorithm for fairness. The crossover function is also held constant. Replacement is then handled by iteratively replacing the lowest performing 60% of the population. This approach means that 20% of the original chromosomes that were selected for breeding are immediately replaced after reproduction, 40% are allowed to reproduce and survive to the next generation, and 40% are not allowed to reproduce and are immediately replaced as the cycle concludes [4].

*c) Hybrid Selection*

In the hybrid selection strategy, the age-based algorithm described earlier is enhanced with elitism to allow the fittest chromosome to breed with the next fittest member of the population and to survive unchanged into the next generation. This approach ensures that the most optimum knowledge of the population is never selectively edited out of the population. As the offspring of the elite member cannot replace the elite parent, it overwrites the offspring of the lowest performing member instead [4].

### D. Senescent Genetic Algorithms

This group of experimental algorithms is based upon a unique group of evolutionary aging theories. Rapid Senescence, Gradual Senescence, and Non-Linear Senescence will all be compared directly to the performance of the conventional genetic algorithms, whereas the CA+GA will mainly be compared to itself using both aging and non-aging chromosomes.

*a) Rapid Senescence*

In rapid senescence, each chromosome object has the additional attributes "max age" and "current age". In this algorithm, selection is handled similarly to the fitness-based algorithm with the addition of an aging function. This aging function works by comparing each chromosome's current age (a counter of how many generations the chromosome has survived) to the maximum age. The maximum age for this experiment was set at 25 which was experimentally determined (discussed later) to be the most optimum age. If the current age exceeds the maximum age, the chromosome is placed in the lowest performing 40% of the population regardless of its fitness so that it may be replaced. This mechanism for handling the senescent death of chromosomes adds the possibility for the entire 40% of the top performing chromosomes to be replaced concurrently if they were to meet their maximum age at the same time [7].

*b) Gradual Senescence*

Gradual senescence is handled very similarly to rapid senescence with the exception that gradual senescence does not have a maximum age for which an individual may live. Instead, gradual senescence employs an aging function that edits the fitness of a chromosome to make the fitness gradually decline as the chromosome survives more generations. The aging function was defined as a cubic function [12] so that aging would have a minimal effect early in life but would become exponentially more effective as the chromosomes survive for longer periods. The aging function was experimentally determined and defined as follows:

*Equation 1*

$$\text{Let } C = Current\ age\ of\ a\ chromosome$$
$$\text{Let } D = Distance\ of\ a\ chromosome's\ trip$$
$$\text{Let } F = Calculated\ fitness\ of\ a\ chromsome$$

$$F = D + C^3/1000$$

As a lower fitness is selected for, this approach puts the selective pressure more on editing out chromosomes that have survived longer and allows newer chromosomes a greater chance at survival. If a chromosome happens to be of a much higher fitness than all its competitors though, it can continue to survive even with the effects of the aging function acting against it in "old age" [8].

*c) Non-linear Senescence*

In non-linear senescence, there is no set number of chromosomes that will be selected to breed. Instead, each chromosome is assigned a "stage" at each iteration of reproduction, growth, or repair. In this model, there is an unequal favor towards a chromosome being in the reproduction stage as compared to the other two stages. The chance that a chromosome will be in reproduction is set at 50% while the other stages both comprise 25%. This weight mimics the Disposable Soma Theory [13] in that individuals focus their limited resources towards reproduction rather than growth or repairing their cells. Chromosomes are all assigned a starting "age" of 52 (experimentally determined later). From there, the chromosomes may use their resources towards being in one of the three possible states. The states are defined to affect the chromosome as follows:

- Reproduction: subtracts 0.7 generations from the chromosome's life.
- Growth: subtracts 0.3 generations from the chromosome's life.
- Repair: adds 0.6 generations to the chromosome's life.

These numbers ensure that if all three states are selected exactly according to their assigned weights, this would lead to an average maximum age of around 141 generations as every three generations, a chromosome will tend to lose 1.1 generations of lifespan. This technique is most like rapid senescence in that it completely removes the implementation of aging from fitness except for the fact that aging will only affect an individual if that individual is fit enough to meet its maximum lifespan. Once an individual meets its maximum lifespan in this implementation, it is immediately replaced.

*d) Cellular Automata Enhanced Genetic Algorithm*

The CA+GA algorithm is not directly intended to test one of the evolutionary theories as the others are. Instead, it gauges the effect of localization and geographic isolation on death. This algorithm uses the Programmed Death Theory as its basis for controlling aging. Chromosomes are initialized and placed in a 10 x 10 matrix with all edges connected to create a torus [5]. This avoids any effect that edges would have on the population's performance. Each cell in the matrix is then iterated through. Fitness is measured every time a cell is changed as to maintain the correct fitness measure. Each cell can randomly select one of the nine surrounding cells for mating. If the cell is empty (from a previous chromosome dying), then the current cell can reproduce asexually to fill that cell with a guaranteed mutation of one gene. If the cell is populated with another chromosome, then the cells will produce a single offspring with a two-point crossover function and the same chance at mutation as was used for all other experiments. If the offspring is more fit than the least fit parent, then the offspring will assume that parent's position in the matrix. This method covers the selective pressure towards a fitter population. Any new offspring are set to a current age of zero to allow them to go through the full aging process. The maximum age for each chromosome is again set to 25 generations as with the Programmed Death Theory experiment.

In this experiment, 100 iterations are performed on the same problem with both chromosomes whose maximum age is set to 25 generations and with chromosomes whose maximum age is set beyond the total number of generations for which the algorithm is allowed to run. Essentially, this makes them immortal in terms of aging and allows us to directly view the effects of senescence in a spatial environment with local reproduction [1, 5, 14].

IV. EXPERIMENTAL RESULTS

*A. Experiment 1: Best Conventional Genetic Algorithm*

Experiment 1 attempted to find the best-performing conventional genetic algorithm to establish a baseline for which to compare the senescence-enhanced algorithms in experiments 2, 3, and 4.

First, the age-based conventional algorithm was run 100 times until the stopping criteria was met. Averaged over each run, the total run-time of the algorithm was 148.36 seconds and the optimal distance found was high at 2,617.76. It tended to take only 2,330.71 generations to find its optimal solution. As there were 20,000 potential generations that it was able to use, that means that typically 17,669.29 generations passed without any progress being made towards a more optimal solution. This is likely due to a failure to maintain knowledge between generations – the algorithm tends too far towards exploration and does not fully exploit the optima it is currently moving towards.

Next, the fitness-based algorithm was run and data collected over all 100 iterations. This algorithm tended to perform much more optimally than the age-based algorithm with an average distance found of just 836.5. The total run-time was much higher at 362.59 seconds but tended to find its optimum solution at 206.19 seconds while utilizing 11,347.28 generations. This higher run-time is caused by the fitness-based algorithm's ability to make progress towards a more optimal solution for a longer period during its run.

Finally, the hybrid algorithm was able to find an optimal distance of 942.83. It took the longest of all three algorithms to run fully at 447.03 seconds. The hybrid approach made use of the longest portion of its run to find an optimal solution at 385.93 seconds and 14,107.32 generations.

Overall, the fitness-based approach was clearly the most optimal of the three conventional approaches tested. It also had the lowest deviation of the three, which means that not only does it find the best solution, but running the algorithm multiple times on the same data set would show that the solutions it provides are less varied, making it a more consistent approach. The fitness-based selection method will be the baseline for which to compare experiments 2, 3, and 4.

### B. Experiment 2: Fitness-based Selection VS Rapid Senescence

Experiment 2 sought to determine whether a death based on theories of rapid senescence could be used to improve a conventional genetic algorithm. As the fitness-based selection approach proved to be the best-performing approach from Experiment 1, this approach was modified to include a programmed death.

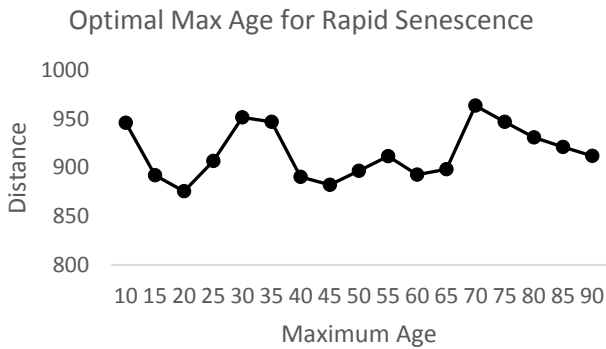

*Figure 1*

Before running all 100 iterations of this algorithm to obtain a statistically significant result, the algorithm was run while incrementally adjusting the maximum age from 10 to 90 in increments of five collecting five samples at each iteration. This was used to get a shallow overview of the general range which would provide the most optimal results. This overview showed optima around maximum ages of 20 and 45 [Figure 1]. More iterations were then pulled around these optima and after these additional runs, the maximum age was set at 25 as it provided the most optimal results.

After 100 iterations with a maximum age of 25, the average optimal distance found by the rapid senescence algorithm was 817.31 making the algorithm 2.29% more optimal than a fitness-based approach. The total run-time was similar to the fitness-based conventional approach at 363.09 seconds (a one second difference). Overall, the rapid senescence algorithm took 15,290.47 generations to find its optimal solution, showing that the algorithm can make progress for a longer period than all the conventional algorithms tested. This demonstrates that death may be an important factor in keeping a population moving towards optimality rather than getting trapped in a local optimum. The number of senescent deaths were also tracked during this experiment to show how many chromosomes died an evolutionary death as compared to a death from becoming unfit by other means. The proportion of senescent deaths in this experiment was 5586.02 senescent deaths to 12,000 total deaths. Therefore, senescent death comprised 46.55% of all death. Nearly half of all deaths were an evolutionary death to provide this optimal performance.

### C. Experiment 3: Fitness-based Selection VS Gradual Senescence

In Experiment 3, we sought to see if a gradual decline in fitness (gradual senescence) might be more beneficial both in comparison to the rapid senescent technique used in Experiment 1 and in comparison to the fitness-based conventional algorithm. Comparably to Experiment 2, the best aging function was experimentally determined by incrementally increasing the divisor (v) in $F = D + C^3/v$. "v" was increased from 400 to 2,000 in increments of 100. Through this process, the optimal value of "v" was determined to be 1,000.

After 100 iterations were run with the aging function described above, the average distance was 834.43. This indicates the algorithm finds an optimum that is only slightly better that the fitness-based approach and about 2.08% less optimal than the rapid senescence approach. However, the gradual senescence approach takes about 14% less time to complete its run than the rapid senescence approach and can find its optimal solution in just 155.1 seconds and 9,841.81 generations. This means that the gradual senescence algorithm only requires 75.22% of the time that the fitness-based approach takes to find its optimal solution. As senescence begins affecting all chromosomes after their very first iteration, it is not possible to tell what percentage of chromosomes died purely from an evolutionary death.

### D. Experiment 4: Fitness-based Selection VS Non-Linear Senescence

In Experiment 4, the optimal starting "life expectancy" was again iteratively determined by running five iterations of the algorithm from a starting age of 14 up to 68 in increments of four. After this testing, the starting life expectancy was set to 52 generations.

This led to an average distance of 831.7 which is again more optimal than the fitness-based approach (although only 0.57% more optimal). The total run-time was 350.29 seconds making it about 13 seconds faster than the rapid senescence algorithm, slower than the gradual senescence algorithm, and about five seconds faster than the fitness-based approach. The optimum solution tended to be found at 293.92 seconds and 16,835.06 generations, using more of its available 20,000 generations than any of the other algorithms. The number of senescent deaths were tracked in this experiment as well, and on average, 2,655.34 of 299,993.6 deaths could be attributed to aging. This shows that 0.88% of deaths are due to senescence in this model making it have much less effect than the rapid senescence evolutionary theories.

*E. Experiment 5: Immortality VS Senescence with Local Reproduction*

Experiment 5 mimics the spatial systems used by Werfel, Ingber, and Bar-Yam and Mitteldorf [1, 5]. Instead of comparing the results of this experiment to the preceding four experiments, this algorithm was run with both "aging" and "non-aging" chromosomes. The matrix was set to a constant 10 x 10 size to minimize the run-time as the time-complexity grows exponentially for this algorithm in relation to the matrix size.

First, the experiment was run while the chromosomes were given a maximum age beyond the maximum iterations for which the algorithm was allowed to run (4,500 generations). This essentially set no aging on the algorithms and made them "immortal". This immortal trial yielded the average optimal distance of 836.53 while run time sat at 328 seconds. The immortal population took 3,908.24 generations to find its optimal solution.

The optimal maximum age was then iteratively determined as in the other experiments on senescence. This experimentation set the maximum age at 45 generations. The effect of this age limit combined with rapid senescence provided an optimal distance of 833.06 (3.47 less than the immortal population). The aging population took an average of 332.61 seconds to complete its run, but it also tended to take one less generation than the aging population to find an optimal solution.

In the context of the aging CA+GA algorithm, the maximum age of 45 leads to a small percentage of senescent death to provide optimal performance. We believe this is since a chromosome's offspring is only able to populate one of the surrounding nine cells if it is more optimal than the chromosome that is currently in that location. This ensures that an offspring has a relatively low fitness and it must have a lower age than the parent itself. This leads to an "inbreeding effect" in which the surrounding cells are likely to become more fit than the parent and replace it before the parent can die a senescent death.

## V. CONCLUSIONS

The results of this experimentation show that evolutionary theories of senescent death can make a significant impact in the performance of genetic algorithms in terms of the optimality of a solution, the consistency of a solution, and the time needed to find it. This data shows a non-trivial improvement in all these areas. Experiment 2 reveals that the greatest improvement in accuracy while improving run-time can be accomplished by using a rapid senescence approach. Experiment 3 establishes the ability of death to evolve a population more quickly. Experiment 4 demonstrates the variability that can be used in controlling a chromosome's aging while still making an improvement over the classical approaches. Experiment 5 illustrates the effects of aging and its ability to optimize performance in a geographically isolated population.

Senescent genetic algorithm enhancement is attractive in that it does not have to be the only improvement made to a conventional genetic algorithm. Senescence can and should be applied to a variety of enhanced genetic algorithm techniques that have been studied and applied over the years.

Overall, these experimental results prove that death is an evolved characteristic. Death may not provide an evolutionary benefit directly to the individual who is experiencing a senescent death at that moment, but, with a more global perspective, that individual altruistically contributes to the greater good of its community by allowing for its own removal from it just as its predecessors had done. In this way, senescent death is not only beneficial for the individuals who remain in the next generation, but it is beneficial to all generations that have ever passed. Death therefore is not just an evolved trait, but it is a driving force for the constant pursuit of the goal of reaching optimality in one's environment. Ultimately, death is not the antagonist of life, but it is a tool that can be used to make life better.